\documentclass[pdflatex,sn-vancouver-num,Numbered]{sn-jnl}

\usepackage{tikz}
\usetikzlibrary{calc,arrows.meta}
\usepackage{array}
\usepackage{graphicx}
\usepackage{multirow}
\usepackage{amsmath,amssymb,amsfonts}
\usepackage{amsthm}
\usepackage{mathrsfs}
\usepackage[title]{appendix}
\usepackage{xcolor}
\usepackage{url}
\usepackage{textcomp}
\usepackage{manyfoot}
\usepackage{booktabs}
\usepackage{hhline}
\usepackage{float}
\usepackage{algorithm}
\usepackage{algorithmicx}
\usepackage{algpseudocode}
\usepackage{listings}

\renewcommand{\orcid}[1]{\href{https://orcid.org/#1}{\textsuperscript{\textsc{orcid}}}}

\begin{document}

\title[Qwen3-VL for HMER]
{Parameter-Efficient Adaptation of Qwen3-VL for
Handwritten Mathematical Expression Recognition}

\author*[1]{\fnm{Manh Tuan} \sur{Ha}\orcid{0009-0004-7583-1148}}
\email{tuanhmse180249@fpt.edu.vn}

\author[1]{\fnm{Xuan Bao Viet} \sur{Tran}\orcid{0009-0000-4975-0601}}
\email{viettxbqe180024@fpt.edu.vn}

\author[1]{\fnm{Vo Minh Dat} \sur{Nguyen}\orcid{0009-0000-7772-8436}}
\email{datnvmqe180070@fpt.edu.vn}

\author[1]{\fnm{Minh Nhat} \sur{Vo}\orcid{0009-0004-9580-581X}}
\email{nhatvmse173616@fpt.edu.vn}

\author[1]{\fnm{Quoc Trung} \sur{Nguyen}\orcid{0000-0002-5500-3554}}
\email{trungnq46@fe.edu.vn}

\author[2]{\fnm{Van Bay} \sur{Nguyen}\orcid{0009-0000-7474-5193}}
\email{bay.nv@ou.edu.vn}

\affil*[1]{%
  \orgdiv{Department of Artificial Intelligence},
  \orgname{FPT University},
  \orgaddress{
    \city{Ho Chi Minh City},
    \country{Vietnam}
  }
}

\affil[2]{%
  \orgdiv{Faculty of Computer Science},
  \orgname{Ho Chi Minh City Open University},
  \orgaddress{
    \city{Ho Chi Minh City},
    \country{Vietnam}
  }
}

\abstract{Handwritten mathematical expression recognition (HMER) requires joint recovery of symbol identity and two-dimensional syntax, so a locally plausible transcription may still be structurally wrong. We study whether a compact vision--language model can be adapted to generate linearized Symbol Label Graphs (SLGs) under limited computing resources. Starting from a 4-bit Qwen3-VL-4B checkpoint, we apply rank-16 Low-Rank Adaptation to the vision, language, attention, and MLP modules and train with instruction-formatted CROHME 2019 examples. We compare this model with an internally implemented ViT--Transformer baseline, a zero-shot Qwen3-VL configuration, and additional internally evaluated alternatives. On the 1,198 valid CROHME test examples, the CROHME-only model obtains 524 exact matches (43.74\%) and a corpus BLEU-4 score of 77.67, compared with 4.92\% and 43.26 for the task-specific ViT baseline. An ablation that adds 2,000 manually written, formula-guided samples decreases exact match to 42.32\%, exposing a domain-alignment issue rather than an automatic benefit from more data. The principal contribution is therefore an evidence-backed, resource-efficient formulation of HMER as multimodal structural generation, together with a controlled data-composition ablation and an analysis of the remaining gap between local token overlap and globally correct structure.}



\maketitle

\section{Introduction}
The reliable translation of handwritten mathematical expressions into standard computable formats such as \LaTeX{} remains a difficult problem in document analysis, computer vision, and educational technology \cite{zanibbi2012recognition}. Unlike ordinary optical character recognition, mathematical expression recognition must recover both symbol identity and two-dimensional layout. A small spatial error can change the meaning of an expression, for example when a digit is interpreted as a normal right-neighbor rather than a superscript. This makes handwritten mathematics particularly challenging because the input combines writer-dependent stroke variation, ambiguous symbol shapes, and structural relations such as fractions, roots, subscripts, superscripts, and nested expressions.

Early work on handwritten mathematical expression recognition emphasized structural analysis and graph-based representations, while later systems increasingly adopted neural encoder-decoder models that translate expression images directly into markup sequences \cite{deng2017image,zhang2017watch,wang2020multiscale}. The CROHME competitions have provided a common benchmark for evaluating these methods and have highlighted the persistent gap between visual recognition and structural interpretation \cite{mouchere2014icfhr,mahdavi2019icdar}. Recent transformer-based approaches use bidirectional training, counting-aware modeling, syntax-aware decoding, and coverage modeling \cite{zhao2021bttr,li2022counting,yuan2022syntax,zhao2022comer}. Peer-reviewed work on vision transformers, transformer OCR, and document understanding also demonstrates the value of pretrained visual representations for structured recognition \cite{dosovitskiy2021image,liu2021swin,li2023trocr,kim2022donut,huang2022layoutlmv3}. These developments motivate a closer investigation of whether a modern multimodal model can be adapted to handwritten mathematical expressions when the output is not plain text but a structured label graph.

This study formulates HMER as instruction-conditioned generation of a linearized Symbol Label Graph (SLG). The formulation retains explicit relations between symbols while allowing a pretrained vision--language decoder to be adapted with Low-Rank Adaptation (LoRA) and 4-bit quantized weights \cite{hu2022lora,dettmers2023qlora}. We deliberately avoid claiming state of the art: the goal is a reproducible, resource-efficient study of structural adaptation, evaluated under one consistent local protocol.

The contributions are fourfold:
\begin{itemize}
    \item We adapt a 4-bit, 4B-parameter vision--language checkpoint to direct image-to-SLG generation by applying rank-16 adapters across both visual and language components.
    \item We provide a controlled comparison with an internally implemented ViT--Transformer encoder--decoder and show that multimodal pretraining primarily improves global sequence consistency, not merely token overlap.
    \item We report a data-composition ablation using 2,000 formula-guided handwritten samples. Its negative result identifies distribution alignment as a more important issue than raw sample count.
    \item We distinguish strict exact match from BLEU-4 and analyze why strong local overlap does not guarantee a completely correct mathematical structure.
\end{itemize}

\section{Related Work}
Handwritten mathematical expression recognition has traditionally been studied as a combination of symbol segmentation, symbol classification, and structural relation analysis \cite{zanibbi2012recognition}. In this setting, the model must determine which strokes belong to the same symbol, identify the symbol class, and infer the spatial grammar connecting the symbols. This differs from ordinary handwriting recognition because the reading order is not always horizontal and because mathematical notation encodes meaning through two-dimensional placement. A fraction bar, a radical sign, or a superscript can change the semantic interpretation of the entire expression even when the individual symbols are recognized correctly.

The CROHME benchmark has played an important role in standardizing evaluation for this task. Earlier CROHME competitions encouraged methods based on graph grammars, stroke grouping, and relation classifiers, while later competitions increasingly reflected the transition toward neural sequence modeling \cite{mouchere2014icfhr,mahdavi2019icdar}. The benchmark remains difficult because it contains a mixture of short expressions, long algebraic expressions, nested structures, and writing styles from multiple users. As a result, exact-match accuracy is usually much lower than token-level recognition scores: a prediction may contain many correct symbols but still fail if one spatial relation is wrong.

Neural image-to-markup methods simplified the pipeline by translating an expression image directly into a markup sequence. IM2TEX showed that an encoder-decoder model with attention can generate \LaTeX{} from rendered mathematical images \cite{deng2017image}. Later work extended this direction to handwritten data with attention-based parsing, multi-scale visual features, dense encoders, and transformer decoders \cite{zhang2017watch,wang2020multiscale,zhao2021bttr}. These methods are strong because they allow the model to learn visual and structural cues jointly, but they still require substantial task-specific training and can be sensitive to long-range dependencies. Counting-aware, syntax-aware, contrastive, and coverage-based strategies help maintain consistency between visual evidence and generated structure \cite{li2022counting,yuan2022syntax,lin2022cclsl,zhao2022comer}.

Vision-language models offer a different path. Instead of training a recognition model from scratch, a large multimodal model can be adapted to a specific document understanding task through instruction fine-tuning. Peer-reviewed systems show strong performance in transformer-based text recognition and structured document parsing \cite{li2023trocr,kim2022donut,huang2022layoutlmv3}. However, HMER remains a demanding test case because the output is not a natural language answer but a precise symbolic representation. This study therefore examines whether parameter-efficient fine-tuning can align Qwen3-VL with the CROHME structural format while preserving its general visual reasoning capacity.

\section{Dataset and Preprocessing}

\subsection{Data Source}
The experiments are based on the ICDAR 2019 CROHME dataset for online handwritten mathematical expression recognition \cite{mahdavi2019icdar}. CROHME provides InkML files and ground-truth structural annotations, allowing each expression to be studied as both a handwriting signal and a mathematical structure \cite{w3c2011inkml}. Each InkML file contains time-ordered pen trajectories, while the annotation specifies the symbols and spatial relations that form the target Symbol Label Graph (SLG). This pairing is essential for HMER because correct recognition depends not only on symbol identity, but also on relations such as right-neighbor, superscript, subscript, above, and below.

To test whether additional handwriting styles improve robustness, we also created a supplementary set of 2,000 expressions. Synthetic and graph-conditioned generation has previously been studied as a way to increase formula and handwriting diversity \cite{le2020paired,chen2024generating}. A script generated diverse \LaTeX{} formulas, which were then rendered as visual references and manually rewritten through a custom annotation interface. The interface captured the stroke trajectory and exported each sample in InkML format. This augmented set is used only as an experimental variant; the final results show that the model trained on the original CROHME set remains slightly stronger on the CROHME test split.

\subsection{Preprocessing}
All expressions are converted into a unified SLG-oriented representation. Symbols become graph nodes, and spatial relations become graph edges. The model target is then linearized into a sequence that preserves these graph relations in a \LaTeX-style format. This choice avoids treating a mathematical expression as ordinary left-to-right text and makes the output sensitive to two-dimensional layout. For the multimodal Qwen3-VL setting, each expression image is paired with an instruction prompt, and the expected assistant response is the linearized SLG. For the ViT Hybrid baseline, the same visual input and target sequence are used to ensure a consistent comparison.

\subsection{Problem Formulation}
Let $x$ denote the visual representation of a handwritten mathematical expression and let $y=(y_1,\ldots,y_T)$ denote its target structural sequence. Each token in $y$ can represent either a mathematical symbol or a relation marker derived from the SLG annotation. The recognition task is formulated as conditional sequence generation, where the model estimates $p(y|x)$ and produces the most likely structural transcription. This formulation is flexible enough to support both the ViT Hybrid baseline and the Qwen3-VL model, while keeping the evaluation target identical across methods.

The central difficulty is that local visual correctness is not sufficient. A model may correctly recognize the digits and operators in an expression but still fail exact matching if it places a term in the wrong relation. For example, confusing a right-neighbor relation with a superscript relation changes the mathematical meaning even if both symbols are present. The preprocessing pipeline therefore emphasizes relation-preserving targets. By linearizing the SLG rather than using plain text, the model is encouraged to learn structural dependencies explicitly.

The augmented dataset is treated as an ablation rather than as the default training set. Although generated samples increase writer diversity and help expose the model to additional formula shapes, they may also introduce distributional differences from CROHME. This is especially relevant for HMER, where stroke order, symbol spacing, and writing habits affect the final graph. The evaluation therefore compares CROHME-only fine-tuning against CROHME plus generated data to determine whether additional samples improve or dilute the target-domain signal.

\section{Exploratory Data Analysis}

Table~\ref{tab:data_overview} summarizes the main corpus statistics. The train, validation, and test splits have similar average token lengths, ranging from 18.57 to 19.91 tokens per expression, which suggests that the evaluation split is comparable in difficulty to the training data. The maximum lengths are much larger than the averages, reaching 133 tokens in training and 189 tokens in validation, indicating the presence of structurally complex expressions with long dependency chains.

\begin{table}[htbp]
\centering
\caption{Corpus distribution and token statistics across dataset splits.}\label{tab:data_overview}
\begin{tabular}{@{}lrrrrr@{}}
\toprule
Split & Expressions & Avg. Tokens & Median Tokens & Max Tokens & Min Tokens \\
\midrule
Test  & 1199 & 19.91 & 17 & 91  & 0 \\
Train & 8900 & 18.57 & 15 & 133 & 1 \\
Valid & 986  & 19.32 & 17 & 189 & 1 \\
\botrule
\end{tabular}
\end{table}

The dataset is dominated by common spatial and relational tokens, especially \texttt{Right} and \texttt{NoRel}. This imbalance confirms that HMER is primarily a structural parsing task rather than a simple symbol classification task \cite{zanibbi2012recognition}. The mean stroke count is approximately 14 traces per expression across the three splits, and stroke count correlates positively with target token length. These observations motivate the use of models that can jointly encode visual appearance and long-range structural context.

\section{Model Architecture}

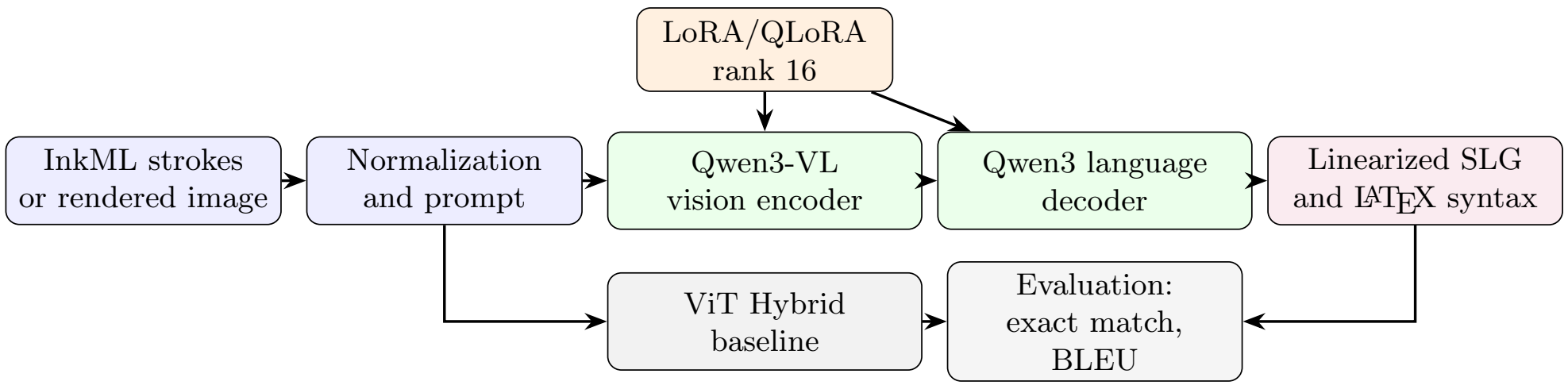
\begin{figure}[H]
\centering
\resizebox{\textwidth}{!}{%
\begin{tikzpicture}[
    >=Stealth,
    font=\small,
    block/.style={draw, rounded corners, align=center, minimum height=9mm, text width=26mm, fill=blue!7},
    model/.style={draw, rounded corners, align=center, minimum height=10mm, text width=30mm, fill=green!8},
    adapter/.style={draw, rounded corners, align=center, minimum height=8mm, text width=24mm, fill=orange!12},
    output/.style={draw, rounded corners, align=center, minimum height=9mm, text width=28mm, fill=purple!8},
    arrow/.style={->, thick}
]
\node[block] (input) at (0,0) {InkML strokes\\or rendered image};
\node[block] (prep) at (3.1,0) {Normalization\\and prompt};
\node[model] (vision) at (6.4,0) {Qwen3-VL\\vision encoder};
\node[adapter] (lora) at (6.4,1.35) {LoRA/QLoRA\\rank 16};
\node[model] (decoder) at (9.8,0) {Qwen3 language\\decoder};
\node[output] (slg) at (13.1,0) {Linearized SLG\\and \LaTeX{} syntax};
\node[model, fill=gray!10] (baseline) at (6.4,-1.45) {ViT Hybrid\\baseline};
\node[output, fill=gray!8] (compare) at (9.8,-1.45) {Evaluation:\\exact match, BLEU};

\draw[arrow] (input) -- (prep);
\draw[arrow] (prep) -- (vision);
\draw[arrow] (vision) -- (decoder);
\draw[arrow] (lora) -- (vision);
\draw[arrow] (lora) -- (decoder);
\draw[arrow] (decoder) -- (slg);
\draw[arrow] (prep) |- (baseline);
\draw[arrow] (baseline) -- (compare);
\draw[arrow] (slg) |- (compare);
\end{tikzpicture}
}
\caption{Overall architecture of the proposed HMER framework. The same preprocessed expression is evaluated with a ViT Hybrid baseline and a LoRA-adapted Qwen3-VL model.}
\label{fig:overall_architecture}
\end{figure}

Figure~\ref{fig:overall_architecture} shows the overall recognition framework. The system receives an online handwritten expression, renders or normalizes it into an image representation, and predicts a linearized Symbol Label Graph. The main model is Qwen3-VL fine-tuned with LoRA/QLoRA adapters, while a ViT Hybrid encoder-decoder is used as a task-specific baseline. This design allows us to compare a conventional vision-to-sequence model against a multimodal foundation model adapted to mathematical handwriting.

\subsection{ViT--Transformer Baseline}
The task-specific baseline is the authors' implementation of a ViT encoder followed by an autoregressive transformer decoder \cite{vaswani2017attention,dosovitskiy2021image}. The code pads grayscale inputs to at most $400\times528$, projects patches to 512-dimensional embeddings, and applies a four-layer, four-head visual encoder. The decoder has six transformer layers, a 512-token vocabulary, and a maximum output length of 256. Training minimizes token cross-entropy with padding ignored. The archived run contains three completed epochs and selects the lowest validation-loss checkpoint. These settings are derived from the submitted code rather than attributed to a pretrained public checkpoint.

This baseline is included for two reasons. First, it provides a controlled comparison with a model whose visual representation is explicitly designed for image classification and sequence decoding. Second, it helps separate the benefit of multimodal pretraining from the benefit of the dataset and preprocessing pipeline. If Qwen3-VL improves substantially over this baseline, the gain can be attributed to both stronger visual-language priors and better adaptation to instruction-style structural generation.

\subsection{Fine-Tuned Qwen3-VL}
The proposed model adapts Qwen3-VL to handwritten mathematical expression recognition. The base checkpoint is \texttt{unsloth/Qwen3-VL-4B-Instruct-bnb4bit}, trained with the Unsloth implementation using LoRA, QLoRA, and 4-bit quantization \cite{hu2022lora,dettmers2023qlora}. Instead of updating all weights, the model trains lightweight rank-16 LoRA adapters on the vision, language, attention, and MLP modules. This reduces the number of trainable parameters and the memory required for task adaptation \cite{hu2022lora,dettmers2023qlora}.

Each example is an instruction-following multimodal conversation: the user turn contains the image and a fixed request for the structural graph, and the assistant turn contains the target SLG. The archived notebook uses a per-device batch size of 2, four gradient-accumulation steps (effective batch size 8), one epoch, a learning rate of $2\times10^{-4}$, five warmup steps, 8-bit AdamW, weight decay 0.01, a linear schedule, and seed 3407. Evaluation and checkpoint saving occur every 20 steps. Two settings are evaluated: CROHME only (8.9K samples) and CROHME plus 2K generated samples (10.9K samples).

\subsection{Parameter-Efficient Adaptation}
Full fine-tuning of a multimodal model is computationally expensive and may overfit when the target dataset is specialized. LoRA addresses this issue by freezing the original model weights and learning a low-rank update for selected projection matrices \cite{hu2022lora}. For a pretrained weight matrix $W_0$, the adapted transformation can be written as:
\begin{equation}
h = W_0x + BAx,
\end{equation}
where $A$ and $B$ are trainable low-rank matrices. In our configuration, the rank is set to 16. This allows the model to learn task-specific visual-to-structure mappings without modifying the full parameter set.

QLoRA further reduces memory usage by combining low-rank adaptation with quantized model weights \cite{dettmers2023qlora}. This is useful for HMER because the task requires multimodal context but the available dataset is much smaller than the general pretraining corpus of a vision-language model. The adapter-based setup acts as a constrained update: it is expressive enough to learn CROHME-specific syntax, yet limited enough to avoid destroying the model's general visual representation.

\subsection{Training Strategy}
The instruction prompt is intentionally simple. It asks the model to read the mathematical expression from the image and output the corresponding structural label graph. The prompt is kept stable across all training examples so that the model focuses on the image-to-structure mapping rather than on prompt variation. During training, the target response is the complete linearized SLG, including relation tokens. This makes the loss sensitive to both symbol recognition and structural ordering.

The two fine-tuning settings test different assumptions about data scale. The CROHME-only setting maximizes domain consistency, while the augmented setting increases sample diversity. Because generated handwriting can differ from the CROHME distribution, especially in stroke spacing and symbol style, the augmented setting is not assumed to be automatically better. Instead, it is evaluated as an ablation that measures whether additional synthetic data improves generalization on the official test set.

\section{Evaluation}

The models are evaluated on the 1,198 test records successfully loaded by the evaluation notebook; one empty record present in the 1,199-row exploratory table is excluded. Decoding is greedy with at most 256 new tokens, and whitespace is normalized before strict string comparison. We report exact-match accuracy and corpus BLEU-4 \cite{papineni2002bleu}. Exact match requires the complete structural sequence to equal the reference, whereas BLEU-4 measures partial $n$-gram overlap. These metrics should not be interpreted as graph isomorphism or rendered-formula equivalence.

\begin{table}[htbp]
\centering
\caption{Main performance comparison and ablation on the CROHME test set.}
\label{tab:model-eval}
\begin{tabular}{@{}llrr@{}}
\toprule
\textbf{Model} & \textbf{Training data} & \textbf{Exact match} & \textbf{BLEU-4} \\
\midrule
ViT Hybrid baseline & CROHME (8.9K) & 4.92\% (59/1198) & 43.26 \\
Qwen3-VL zero-shot & none & 0.00\% (0/1198) & 1.32 \\
Qwen2.5-VL fine-tuned & CROHME (8.9K) & 28.38\% (340/1198) & 63.00 \\
IM2TEX reimplementation \cite{deng2017image} & CROHME (8.9K) & 38.74\% & 50.28 \\
\textbf{Qwen3-VL fine-tuned} & \textbf{CROHME (8.9K)} & \textbf{43.74\% (524/1198)} & \textbf{77.67} \\
Qwen3-VL fine-tuned & CROHME + 2K generated & 42.32\% (507/1198) & 75.46 \\
\botrule
\end{tabular}
\end{table}

Table~\ref{tab:model-eval} shows that task adaptation is decisive. The zero-shot configuration produces no exact match, whereas the fine-tuned configuration reaches 43.74\% exact match and 77.67 BLEU-4. Relative to the ViT baseline, this is a gain of 38.82 percentage points in exact match and 34.41 BLEU points. All non-zero-shot rows are results from the authors' local implementations under the reported protocol; they are not copied leaderboard scores and do not imply superiority over the original publications. The 2K-sample ablation slightly reduces both metrics, suggesting a mismatch between the added handwriting and the CROHME test distribution.

\begin{figure}[!htbp]
\centering
\scriptsize
\renewcommand{\arraystretch}{1.22}
\setlength{\tabcolsep}{2pt}
\begin{tabular}{@{}>{\centering\arraybackslash}p{0.12\textwidth}>{\centering\arraybackslash}p{0.22\textwidth}>{\centering\arraybackslash}p{0.30\textwidth}>{\centering\arraybackslash}p{0.30\textwidth}@{}}
\toprule
\textbf{Input} & \textbf{Reference (compiled)} & \textbf{Zero-shot: compiled + raw} & \textbf{Fine-tuned: compiled + raw} \\
\midrule
\textbf{(a) Sample 1039}\par\vspace{-2pt}
\includegraphics[width=\linewidth,height=0.85cm,keepaspectratio]{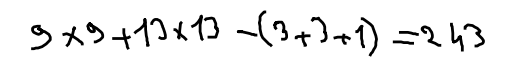}
& $\displaystyle 9\times9+13\times13-(3+3+1)=243$
& $\displaystyle 99+1313-(3+3+1)-243$\par\vspace{1pt}
  {\raggedright\fontsize{6.2}{7.0}\selectfont\ttfamily Raw: 9 9 + 13 13 - ( 3 + 3 + 1 ) - 2 4 3\par}\vspace{1pt}
  {\fontsize{6.5}{7.3}\selectfont\textcolor{red}{No relation tokens; SLG exact: FAIL}}
& $\displaystyle 9\times9+13\times13-(3+{\color{red}\mathbf{7}}+1)=243$\par\vspace{1pt}
  {\raggedright\fontsize{6.2}{7.0}\selectfont\ttfamily Raw: 9 Right \textbackslash{}times Right 9 Right + Right 1 Right 3 Right \textbackslash{}times Right 1 Right 3 Right - Right ( Right 3 Right + Right 7 Right + Right 1 Right ) Right = Right 2 Right 4 Right 3\par}\vspace{1pt}
  {\fontsize{6.5}{7.3}\selectfont\textcolor{red}{Relation-token SLG; exact: FAIL ($3\!\to\!7$)}} \\
\midrule
\textbf{(b) Sample 499}\par\vspace{-2pt}
\includegraphics[width=\linewidth,height=0.85cm,keepaspectratio]{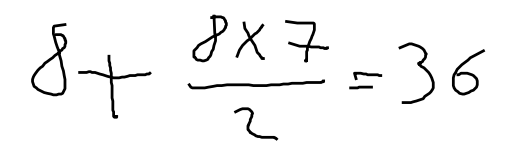}
& $\displaystyle 8+\frac{8\times7}{2}=36$
& $\displaystyle 8+\frac{8\times7}{2}=36$\par\vspace{1pt}
  {\raggedright\fontsize{6.2}{7.0}\selectfont\ttfamily Raw: 8 + (8 * 7) / 2 = 36\par}\vspace{1pt}
  {\fontsize{6.5}{7.3}\selectfont\textcolor{red}{Conventional notation; SLG exact: FAIL}}
& $\displaystyle 8+\frac{8\times7}{2}=36$\par\vspace{1pt}
  {\raggedright\fontsize{6.2}{7.0}\selectfont\ttfamily Raw: 8 Right + NoRel 8 Right \textbackslash{}times Right 7 NoRel - Below 2 NoRel = Right 3 Right 6\par}\vspace{1pt}
  {\fontsize{6.5}{7.3}\selectfont\textcolor{green!45!black}{Relation-token SLG; exact: PASS}} \\
\midrule
\textbf{(c) Sample 614}\par\vspace{-2pt}
\includegraphics[width=\linewidth,height=0.85cm,keepaspectratio]{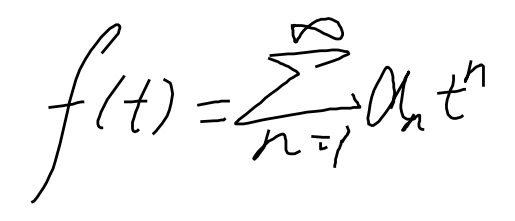}
& $\displaystyle f(t)=\sum_{n=1}^{\infty}a_nt^n$
& $\displaystyle f(t)=\sum_{n=1}^{\infty}a_nt^n$\par\vspace{1pt}
  {\raggedright\fontsize{6.2}{7.0}\selectfont\ttfamily Raw: f(t) = $\Sigma$\_\{n=1\}\textasciicircum{}\{$\infty$\} a\_n t\textasciicircum{}n\par}\vspace{1pt}
  {\fontsize{6.5}{7.3}\selectfont\textcolor{red}{Conventional notation; SLG exact: FAIL}}
& $\displaystyle f(t)=\sum_{n=1}^{\infty}a_nt^n$\par\vspace{1pt}
  {\raggedright\fontsize{6.2}{7.0}\selectfont\ttfamily Raw: f Right ( Right t Right ) Right = Right \textbackslash{}sum Below n Right = Right 1 NoRel \textbackslash{}infty NoRel a Sub n NoRel t Sup n\par}\vspace{1pt}
  {\fontsize{6.5}{7.3}\selectfont\textcolor{green!45!black}{Relation-token SLG; exact: PASS}} \\
\midrule
\textbf{(d) Sample 483}\par\vspace{-2pt}
\includegraphics[width=\linewidth,height=0.85cm,keepaspectratio]{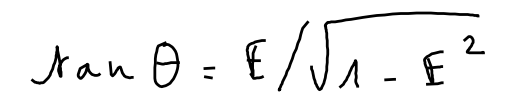}
& $\displaystyle \tan\theta=\frac{E}{\sqrt{1-E^2}}$
& $\displaystyle \tan\theta=\frac{E}{\sqrt{1-E^2}}$\par\vspace{1pt}
  {\raggedright\fontsize{6.2}{7.0}\selectfont\ttfamily Raw: \textasciigrave\textasciigrave\textasciigrave graph; tan$\theta$ = E / $\surd$(1 - E\textsuperscript{2}); \textasciigrave\textasciigrave\textasciigrave\par}\vspace{1pt}
  {\fontsize{6.5}{7.3}\selectfont\textcolor{red}{Fenced conventional notation; SLG exact: FAIL}}
& $\displaystyle \tan\theta=\frac{E}{\sqrt{1-E^2}}$\par\vspace{1pt}
  {\raggedright\fontsize{6.2}{7.0}\selectfont\ttfamily Raw: \textbackslash{}tan Right \textbackslash{}theta Right = Right E Right / Right \textbackslash{}sqrt Inside 1 Right - Right E Sup 2\par}\vspace{1pt}
  {\fontsize{6.5}{7.3}\selectfont\textcolor{green!45!black}{Relation-token SLG; exact: PASS}} \\
\bottomrule
\end{tabular}
\\[4pt]
\parbox{0.96\textwidth}{\scriptsize \textit{Reading note.} Each zero-shot and fine-tuned cell shows the compiled formula first and the model's raw sequence underneath. Panels (b)--(d) have identical compiled formulas, but their raw sequences visibly differ: only fine-tuning produces the required relation tokens. Unicode symbols are rendered with equivalent TeX glyphs for print compatibility.}
\caption{Single-table qualitative comparison for four CROHME test expressions. Panel (a), \texttt{UN19wb\_1111\_em\_1039}, retains the earlier failure case with a $3$-to-$7$ substitution. Panels (b)--(d), \texttt{UN19\_1035\_em\_499}, \texttt{UN19\_1042\_em\_614}, and \texttt{UN19\_1034\_em\_483}, are three additional outputs generated from the saved checkpoint. Although the compiled zero-shot and fine-tuned formulas look identical in these three panels, the raw strings in the same cells show conventional notation for zero-shot and exact relation-token reference SLGs for fine-tuning.}
\label{fig:qualitative_results}
\end{figure}

Figure~\ref{fig:qualitative_results} shows that visually plausible conventional notation can still fail strict SLG scoring. With the same prompt and decoding protocol, fine-tuning supplies the required relation tokens and exactly matches three of the four displayed references; these illustrative cases do not alter the aggregate results.

\begin{table}[htbp]
\centering
\caption{Additional output-level comparison of the Qwen3-VL variants on the same 1,198 test expressions.}
\label{tab:qwen-output-details}
\footnotesize
\setlength{\tabcolsep}{3pt}
\begin{tabular}{@{}lrrrrr@{}}
\toprule
\textbf{Qwen3-VL setting} & \textbf{Exact/non-exact} & \textbf{BLEU-1} & \textbf{BLEU-2} & \textbf{BLEU-3} & \textbf{BLEU-4} \\
\midrule
Zero-shot & 0/1198 & 12.02 & 2.29 & 1.52 & 1.32 \\
Fine-tuned, CROHME & 524/674 & 89.27 & 85.36 & 81.43 & 77.67 \\
Fine-tuned, CROHME + 2K & 507/691 & 87.56 & 82.79 & 78.96 & 75.46 \\
\botrule
\end{tabular}
\end{table}

Table~\ref{tab:qwen-output-details} shows that fine-tuning improves every reported $n$-gram order and raises exact outputs from 0 to 524. The augmented model produces 17 fewer exact outputs and slightly lower BLEU at every order.

\subsection{Discussion}
The gap between 77.67 BLEU-4 and 43.74\% exact match indicates that many residual errors are small but structurally decisive: one incorrect relation token invalidates the entire sequence. The zero-shot result further shows that general multimodal ability is insufficient without supervision for the CROHME grammar. Compared with the ViT Hybrid baseline, the adapted model benefits from pretrained visual-language representations and a decoder already suited to structured generation.

Adding 2K generated samples reduces exact match from 43.74\% to 42.32\% and BLEU-4 from 77.67 to 75.46. The added diversity therefore does not compensate for its apparent mismatch in handwriting dynamics, spacing, or symbol proportions; future augmentation should use filtering, style matching, or curriculum scheduling.

\subsection{Error Analysis}
Failures concentrate in spatial-relation prediction and long sequences. Compact superscripts, adjacent operators, and writer-dependent baselines make right-neighbor and superscript relations visually similar, while an early error in a long sequence can shift all subsequent tokens. Constrained decoding, grammar-aware post-processing, or graph validation may reduce these structurally invalid outputs.

In Fig.~\ref{fig:qualitative_results}(a), fine-tuning recovers the relations but reads one $3$ as $7$; panels (b)--(d) instead show semantically readable zero-shot notation that is invalid for CROHME scoring. Together they distinguish symbol errors from output-grammar errors.

\subsection{Limitations}
The study uses one CROHME 2019 split, one principal seed, and limited configurations, so it does not establish statistical significance or state-of-the-art performance. The IM2TEX and Qwen2.5-VL rows are local implementations, BLEU is not the official CROHME graph metric, and the examples are not a complete error taxonomy.

\section{Conclusion}

This paper investigated the use of Qwen3-VL for handwritten mathematical expression recognition under the CROHME 2019 setting. The proposed framework converts handwriting inputs into linearized Symbol Label Graphs and fine-tunes Qwen3-VL with LoRA/QLoRA adapters. Compared with the zero-shot model and the ViT Hybrid baseline, the fine-tuned Qwen3-VL model shows a substantial improvement in both exact-match accuracy and BLEU-4.

The best configuration is Qwen3-VL fine-tuned on the original CROHME 8.9K training set, achieving 43.74\% exact-match accuracy and a BLEU-4 score of 77.67. The result confirms that domain-specific instruction tuning is essential for aligning a general vision-language model with the structural output format required by HMER. The ablation with 2K generated samples shows a slight performance drop, indicating that synthetic data must be carefully matched to the target handwriting distribution.

The strongest evidence is not simply that a larger pretrained model scores higher. Rather, a compact 4-bit model can be aligned to an exact structural grammar with lightweight adapters, and global structural correctness improves far more than in the from-scratch baseline. At the same time, adding more handwriting is not automatically beneficial: style and annotation alignment matter. Future work should add multi-seed evaluation, the official CROHME scoring tools, grammar-constrained decoding, and similarity-based filtering or curriculum scheduling for the supplementary data.


\bmhead{Acknowledgements}
The authors thank FPT University for academic guidance and computational support. Generative AI tools were used only to assist with language polishing, organization of manuscript text, and \LaTeX{} editing. The authors reviewed and validated the complete manuscript and remain responsible for its accuracy, originality, citations, experimental design, model training, quantitative results, interpretation, and final submission.

\bmhead{Figure rights and permissions}
The architecture diagram, tables, plots, and composite layouts in this manuscript were created by the authors from their own experimental results. The four handwritten-expression inputs in Fig.~\ref{fig:qualitative_results} are research samples from the CROHME 2019 dataset cited in this manuscript and are reproduced for scholarly analysis under the dataset's academic and research terms. No generative-AI images are used. The corresponding author is responsible for retaining any dataset licence or permission documentation requested by the publisher.

\begingroup
\bibliography{ref_main}
\endgroup

\end{document}